\documentclass[journal]{IEEEtran}
\usepackage[T1]{fontenc}% optional T1 font encoding
\usepackage{graphicx}
\usepackage{times}
\usepackage{helvet}
\usepackage{courier}
\usepackage{amsmath}
\usepackage{algorithm}
\usepackage{algorithmic}
\usepackage{csquotes} 
\usepackage{color}
\usepackage{paralist}
\usepackage{amssymb}
\usepackage{indentfirst}
\usepackage{subfigure}
\usepackage{float}
\usepackage{multirow}
\usepackage{cite}
\usepackage{mathrsfs}

\usepackage{mathrsfs} 
\usepackage{amsfonts}
\usepackage{todonotes}
\usepackage{pgfplots} %引用包
\usepackage{epstopdf}
\usepackage{epsfig}
\pgfplotsset{compat=newest}
\usepackage{color}
\definecolor{forestgreen}{RGB}{0,139,69}
\usepackage{CJKutf8}

\usepackage{xcolor}
\definecolor{citecolor}{HTML}{0071bc}
\usepackage[colorlinks, linkcolor=red,  anchorcolor=blue, citecolor=citecolor]{hyperref} 

\usepackage{xcolor}
\definecolor{SeaGreen4}{RGB}{0,205,102} 
\definecolor{SlateBlue}{RGB}{106,90,205} 
\definecolor{DarkRed}{RGB}{178,34,34} 
	
\usepackage[switch]{lineno}
\usepackage{textcomp,booktabs}
\usepackage{amssymb}% http://ctan.org/pkg/amssymb
\usepackage{pifont}% http://ctan.org/pkg/pifont

\usepackage{makecell}
\usepackage{colortbl}
\definecolor{mygray}{gray}{.9}
\definecolor{mypink}{rgb}{.99,.91,.95}
\definecolor{mycyan}{cmyk}{.3,0,0,0}

\begin{document}

\title{  EventSTR: A Benchmark Dataset and Baselines for Event Stream based Scene Text Recognition  }

\author{Xiao Wang, \emph{Member, IEEE}, Jingtao Jiang, Dong Li, Futian Wang*, Lin Zhu, \\ 
    Yaowei Wang, \emph{Member, IEEE}, Yongyong Tian, \emph{Fellow, IEEE}, Jin Tang 

\thanks{Xiao Wang, Jingtao Jiang, Dong Li, Futian Wang, and Jin Tang are with the School of Computer Science and Technology, Anhui University, Hefei 230601, China. (email: \{xiaowang, wft, tangjin\}@ahu.edu.cn)} 

\thanks{Lin Zhu is with Beijing Institute of Technology, Beijing, China (email: linzhu@pku.edu.cn)} 

\thanks{Yaowei Wang is with Peng Cheng Laboratory, Shenzhen, China; Harbin Institute of Technology (HITSZ), Shenzhen, China. (email: wangyw@pcl.ac.cn) }  

\thanks{Yongyong Tian is with Peng Cheng Laboratory, Shenzhen, China; National Key Laboratory for Multimedia Information Processing, School of Computer Science, Peking University, China; School of Electronic and Computer Engineering, Shenzhen Graduate School, Peking University, China. (email: yhtian@pku.edu.cn)}

\thanks{* Corresponding author: Futian Wang (email: wft@ahu.edu.cn)}  
}

\markboth{ IEEE Transactions on ***, 2025 } 
{Shell \MakeLowercase{\textit{et al.}}: Bare Demo of IEEEtran.cls for IEEE Journals}

% make the title area
\maketitle

% As a general rule, do not put math, special symbols or citations in the abstract or keywords.
\begin{abstract}
Mainstream Scene Text Recognition (STR) algorithms are developed based on RGB cameras which are sensitive to challenging factors such as low illumination, motion blur, and cluttered backgrounds. In this paper, we propose to recognize the scene text using bio-inspired event cameras by collecting and annotating a large-scale benchmark dataset, termed EventSTR. It contains \textit{9,928} high-definition ($1280 \times 720$) event samples and involves both Chinese and English characters. We also benchmark multiple STR algorithms as the baselines for future works to compare. In addition, we propose a new event-based scene text recognition framework, termed SimC-ESTR. It first extracts the event features using a visual encoder and projects them into tokens using a Q-former module. More importantly, we propose to augment the vision tokens based on a memory mechanism before feeding into the large language models. A similarity-based error correction mechanism is embedded within the large language model to correct potential minor errors fundamentally based on contextual information. Extensive experiments on the newly proposed EventSTR dataset and two simulation STR datasets fully demonstrate the effectiveness of our proposed model. We believe that the dataset and algorithmic model can innovatively propose an event-based STR task and are expected to accelerate the application of event cameras in various industries. The source code and pre-trained models will be released on \url{https://github.com/Event-AHU/EventSTR}. 
\end{abstract}


\IEEEpeerreviewmaketitle

\section{Introduction}

\begin{figure*}
\centering
\includegraphics[width=\linewidth]{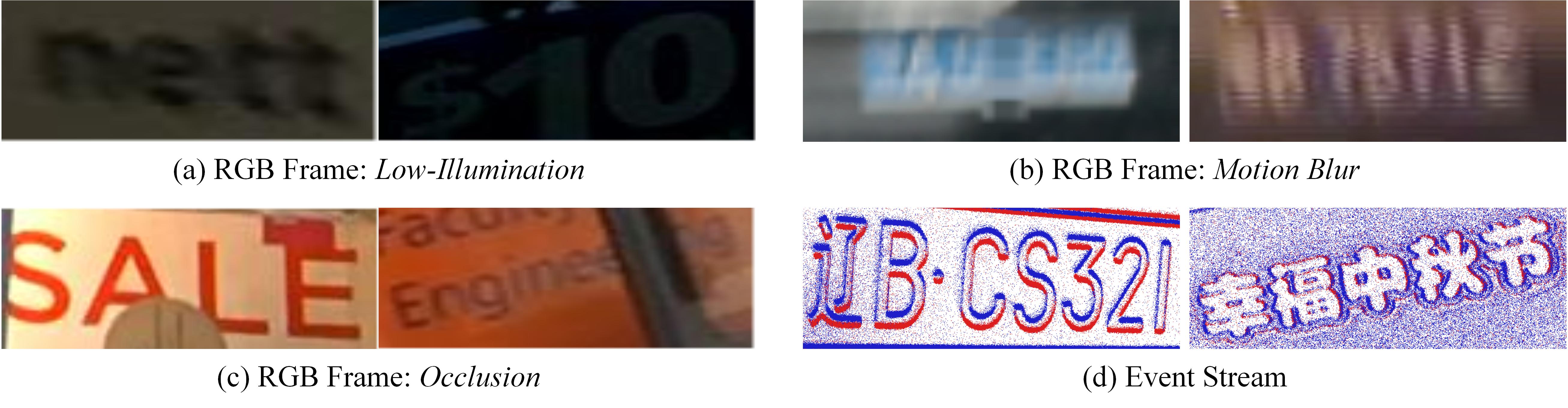}
\caption{Examples illustrating the motivation behind EventSTR. (a) Challenges of scene text recognition under low-light conditions where RGB cameras struggle to capture clear text. (b) Motion blur scenarios that degrade text readability in RGB images. (c) Occlusion issues that hinder text recognition in complex environments. In contrast, (d) shows event camera data that effectively addresses low-light and motion blur challenges due to its high temporal resolution and dynamic range. Additionally, occlusion issues can be mitigated through the reasoning capabilities of LLMs, enabling more robust text recognition in challenging scenarios.} 
\label{fig:motivation}
\end{figure*}

%% background 
\IEEEPARstart{S}{cene} Text Recognition (STR), often referred to as Optical Character Recognition (OCR) in the context of images or videos, is the process of detecting and recognizing text that appears in real-world photographs taken from arbitrary viewpoints and conditions. This technology enables machines to ``read” text within scenes, which can be valuable for various applications such as data entry automation, translating documents, and enhancing the accessibility of digital content. Usually, the STR model is developed for RGB cameras which suffer from low illumination, cluttered backgrounds, fast motion, etc, as shown in Fig.~\ref{fig:motivation} (a-c).

The performance of scene text recognition has been boosted significantly in the deep learning era with the release of large-scale RGB images or synthetic data. For example, methods like PARSeq~\cite{bautista2022PARseq} and MGP-STR~\cite{wang2022mgpstr} utilize attention mechanisms to model character sequences effectively, achieving high accuracy on benchmark datasets. Inspired by the success of the large models in natural language processing, some researchers also exploit to recognize the scene text using large vision-language models. Specifically, methods like mPLUG-DocOwl~\cite{hu2024mplug}, TextMonkey~\cite{liu2024textmonkey}, and DocPedia~\cite{feng2023docpedia} leverage large vision-language models for scene text recognition, enhancing text-image interaction and document understanding. However, the inference in the practical scenarios is still unsatisfied due to the aforementioned issues.

Recently, event cameras draws more and more attention in the computer vision community. Many researchers attempt to introduce event cameras to help or even replace the RGB cameras for their tasks, such as object detection and tracking~\cite{li2022asyDET}~\cite{wang2024eventvot}~\cite{wang2024mambaevt}, pattern recognition~\cite{li2023semantic}~\cite{wang2024hardvs}, semantic segmentation~\cite{jia2023eventSEG}, etc. Many works demonstrate the effectiveness and advantages of event cameras on low power consumption, low latency, high temporal resolution, and high dynamic range. The fundamental reason lies in that the event cameras emit a spike/event point $(x, y, t, p)$ only when the variation of corresponding pixels is beyond the certainty threshold. Here, $(x, y)$ denotes the spatial coordinates, $t$ is the timestamp and $p$ denotes polarity (i.e., positive or negative event). The scene text recorded using an event camera is visualized in Fig.~\ref{fig:motivation} (d).

\begin{figure*}[!htp]
\centering
\includegraphics[width=\linewidth]{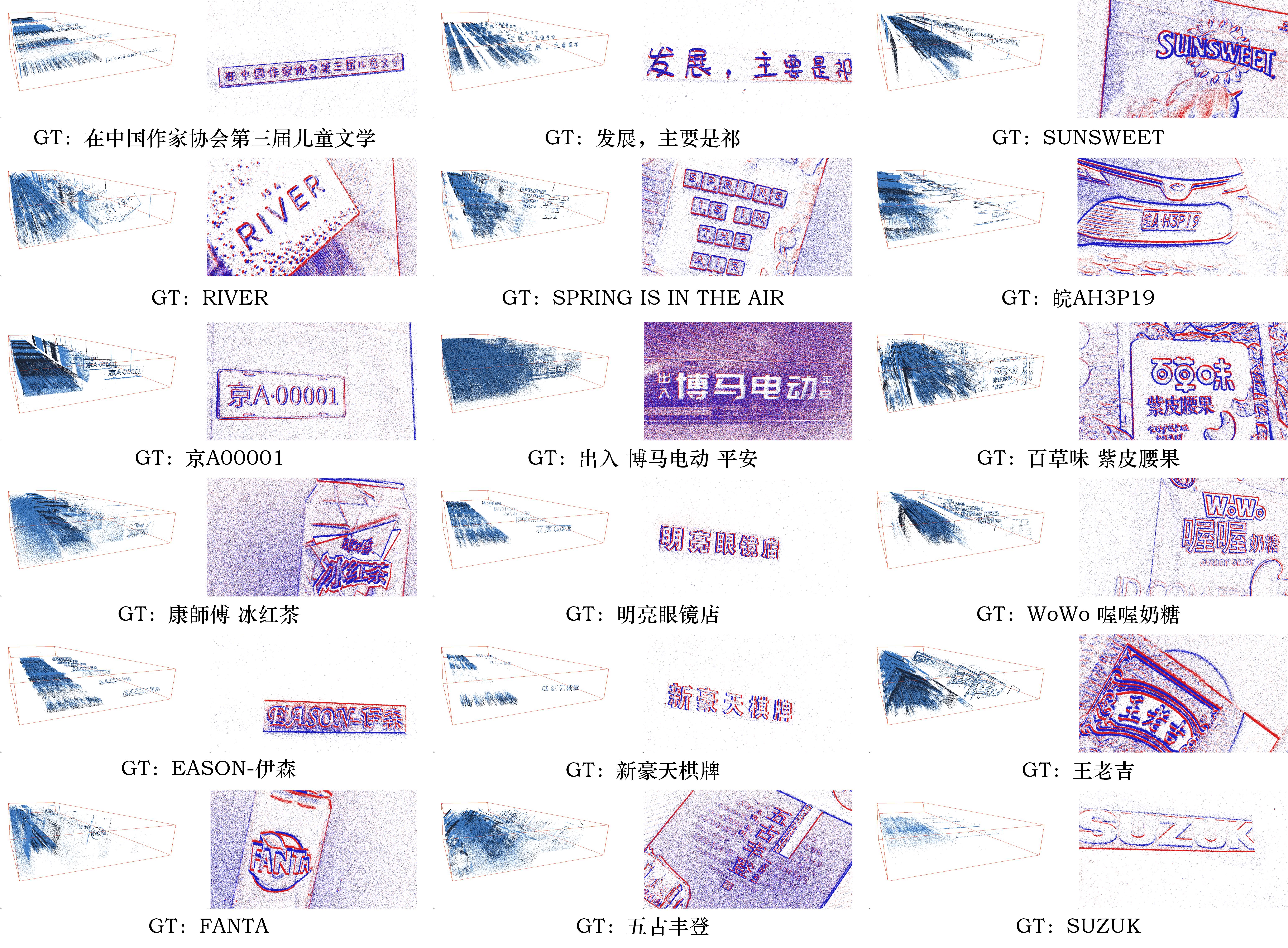}
\caption{Illustration of some representative samples of our proposed EventSTR dataset. 
The left side displays the event stream, while the right side shows the corresponding first frame image.}
\label{fig:dataset}
\end{figure*}

Considering the features and advantages of event cameras for perception, in this paper, we formally propose event stream based scene text recognition by providing a large-scale benchmark dataset and a large language model based event STR framework. The event-based STR dataset \textbf{EventSTR} contains \textit{9,928} samples that fully reflect the key features of event cameras. More in detail, these event samples are collected under different lighting conditions, motions, occlusions, scene categories, and text orientations. Also, the collected data has a resolution of $1280 \times 720$, which can effectively support research on processing high-resolution neural networks. In addition, we also provide multiple baselines for this dataset which will be useful for future works to compare. Some representative samples of the EventSTR are visualized in Fig.~\ref{fig:dataset}.

On the basis of the newly proposed dataset, we further propose a new baseline approach for event-based scene text recognition, termed \textbf{SimC-ESTR}. Specifically, given the event stream, we first obtain its feature embeddings using a vision backbone network, meanwhile, we also adopt the Q-former module to transform the vision features to better adapt to the large language model. The vision features obtained from simple projection and Q-former modules are fed into the pre-trained large language model (LLM) together with generation prompts. We also introduce the memory mechanism to augment the visual features based on context samples. 
\begin{CJK*}{UTF8}{gbsn} 
Additionally, we find that text recognition in scenarios involving Chinese characters is often susceptible to interference from visually similar characters. For the character ``枫'', its visually similar characters include ``松'', ``柏'', ``柳'', and ``杨''. Therefore, we design a new similar word database to help the LLM refine the generated text due to the homophonic Chinese characters. 
\end{CJK*}
Extensive experiments on three benchmark datasets fully validated the effectiveness of our proposed modules for the large language model based event scene text recognition.

To sum up, we draw the contributions of this paper as the following three aspects: 

1). We propose for the first time the task of scene text recognition based on event cameras, aiming to address challenging factors such as low illumination, complex backgrounds, and motion blur. To support this, we have constructed a large-scale high-definition event stream scene text recognition database, termed {EventSTR}. 

2). We have developed a framework for event stream scene text recognition based on large language models, termed {SimC-ESTR}. It simultaneously incorporates a memory mechanism for contextual sample augmentation and visually similar word error correction, achieving superior recognition accuracy.

3). We provide an extensive benchmark involving multiple state-of-the-art scene text recognition algorithms. Additionally, we conducted detailed experimental analyses on three datasets, fully verifying the effectiveness of the proposed method.

\textbf{\textit{The rest of this paper is organized as follows:}} 
In Section~\ref{sec::relatedWorks}, we give a review of the related works including scene text recognition, large language model based OCR, and event-based vision. Then, we propose the key frameworks in Section~\ref{sec::method} by providing the overview, detailed network architectures, and loss functions. In Section~\ref{sec::EventSTR-benchmark}, we propose the EventSTR benchmark dataset with a focus on the protocols, data collection and annotation, statistical analysis, and benchmark baselines. The experiments are conducted in Section~\ref{sec::Exp} and we conclude this paper in Section~\ref{sec::conclusion}.

\section{Related Works} \label{sec::relatedWorks}

In this section, we review the related works on Scene Text Recognition, LLM-based OCR, and Event-based Vision. More related works can be found in the following surveys~\cite{long2021STRDsurvey, wang2023MMPTMsurvey, gallego2020eventsurvey} and paper list~\footnote{\url{https://github.com/Event-AHU/Event_Camera_in_Top_Conference}}.

\subsection{Scene Text Recognition} 
Scene text recognition~\cite{wang2011end,shi2016end,liao2019scene,han2024spotlight} naturally involves both vision and language processing. 
% However, much of the previous research has concentrated on enhancing either visual feature extraction or language modeling independently. The challenge lies in effectively integrating these two modalities to reduce the reliance on one over the other. Recent approaches have sought to balance the contributions of both modalities, achieving a more robust and accurate recognition system across diverse scenarios. These methods aim to optimize the joint contributions of vision and language features, improving performance in varying contexts.
E2STR~\cite{zhao2024E2STR} enhances adaptability to diverse scenarios by introducing context-rich text sequences and applying a context training strategy, enabling flexibility in recognizing texts across different environments. 
% However, its complex inference process increases computational overhead, making it less suitable for real-time applications or resource-constrained environments. 
Guan et al. propose CCD~\cite{Guan2023CCD}, a self-supervised character-to-character distillation method, which learns robust text feature representations via a self-supervised segmentation module and flexible augmentation techniques. 
% While CCD improves recognition, it still struggles with extreme conditions like blurred or occluded text. 
SIGA~\cite{guan2023SIGA} optimizes self-supervised segmentation and implicit attention alignment to improve attention accuracy, though it is constrained when character-level annotations are insufficient. 
CDistNet~\cite{zheng2024cdistnet} incorporates visual and semantic positional embeddings into its transformer-based architecture, offering improvements but still facing difficulties with irregular or dense text layouts and complex backgrounds, limiting its generalization capacity. 
% PARSeq~\cite{bautista2022PARseq} presents a novel approach to scene text recognition by learning a shared-weight autoregressive language model ensemble through parallel language modeling. 
% % It unifies autoregressive and non-autoregressive inference while using bidirectional context for iterative optimization, enhancing recognition accuracy. However, this approach also introduces additional complexity in training and inference, which could lead to inefficiencies in deployment, particularly in real-time scenarios. 
% TPS++~\cite{zheng2023tps++} introduces an attention-enhanced transformation for scene text recognition, improving on traditional TPS by using content-aware correction to handle highly distorted text.
In contrast, another group of approaches focuses on using language models for iterative error correction in scene text recognition. These methods refine recognition results by correcting errors during inference, resulting in more robust and interpretable systems. Recent models, such as VOLTER~\cite{li2024volter}, BUSNet~\cite{Wei2024BUSNet}, MATRNet~\cite{na2022matrn}, LevOCR~\cite{da2022levocr}, and ABINet~\cite{fang2021ABINet}, integrate language models for this iterative correction. 
Inspired by these models, in this paper, we also exploit large language model based scene text recognition using an event camera.

\subsection{LLM-based OCR} 
Recent advancements in scene text recognition have harnessed the power of large language models (LLMs)~\cite{zhao2023survey} to enhance text understanding and improve recognition accuracy. These models focus on optimizing the integration between visual features and linguistic context, offering significant improvements over traditional methods. TextMonkey~\cite{liu2024textmonkey} is a multimodal LLM optimized for text-centric tasks, providing enhanced interaction and interpretability through high-resolution inputs and location-aware responses. 
% With improved image feature filtering and extended task capabilities, TextMonkey demonstrates strong potential in downstream applications, including interactive app navigation. 
DocPedia~\cite{feng2023docpedia} is an advanced multimodal model designed for OCR-free document understanding, capable of processing high-resolution images directly in the frequency domain to capture both visual and textual information efficiently. 
Vary~\cite{wei2025vary} enhances the visual vocabulary of large vision-language models (LVLMs), specifically designed for tasks that require dense and fine-grained visual perception, such as document-level OCR and chart interpretation. 
% It excels in improving fine-grained perception and understanding, particularly for non-English contexts. 
mPLUG-DocOwl 1.5~\cite{hu2024mplug} introduces Unified Structure Learning, improving text-rich document image understanding in multimodal large language models (MLLMs). 
% Fox~\cite{liu2024focus} presents an effective pipeline and hybrid data strategy to enhance the fine-grained understanding of multi-page documents in large language models. 
% By focusing on document-level regions and integrating multi-visual vocabularies, it supports format- and page-agnostic comprehension. UReader~\cite{ye2023ureader} is a cost-effective, OCR-free visual language understanding model based on a multimodal LLM. It fine-tunes only 1.2\% of the parameters while achieving top performance in various visual language understanding tasks through joint fine-tuning and auxiliary tasks, without requiring downstream task fine-tuning. 
OCR2.0~\cite{wei2024OCR2.0} introduces an advanced end-to-end model with 580M parameters, leveraging large language models (LLMs) to handle a wide range of OCR tasks, from text to formulas and diagrams. 
% Its interactive OCR features, along with applications in dynamic resolution and multi-page OCR, highlight the power of LLMs in enhancing OCR performance.
However, each of these models has limitations when confronted with extreme conditions, such as poor lighting, low-resolution images, or complex noise. Under these challenging circumstances, maintaining high performance can become difficult.

\subsection{Event-based Vision}  
An event camera~\cite{gallego2020event}~\cite{jiang2024evcslr} is a vision sensor that captures dynamic scenes with microsecond-level time resolution by recording pixel-level brightness changes rather than fixed-frame images. 
% This enables exceptional performance in high-speed motion, low-light environments, and high-dynamic-range scenes. Event cameras offer low power consumption, high temporal resolution, and flexible responsiveness, making them widely used in fields such as autonomous driving, robotic vision, and medical imaging. 
In human activity recognition, ESTF~\cite{wang2024hardvs} leverages event camera data to capture high-speed and low-light motion by projecting event streams into spatial and temporal embeddings. 
% Using Transformer networks for feature fusion, ESTF enhances activity recognition accuracy in dynamic and challenging environments. 
For object tracking, EventVOT~\cite{wang2024eventvot} introduces the first large-scale high-resolution (1280$\times$720) event-based tracking dataset, containing 1141 videos across multiple categories such as pedestrians, vehicles, drones, and ping pong balls. 
% The paper proposes a novel hierarchical knowledge distillation framework, leveraging multimodal/multiview information during training to enable high-speed, low-latency event-only tracking during testing, addressing the limitations of existing tracking methods that either rely on aligned RGB and event data or suffer from noise and sparse resolution. 
Recurrent Vision Transformers (RVTs)~\cite{gehrig2023RVTs} leverage event cameras' strengths in capturing high temporal resolution and handling challenging lighting conditions to achieve robust detection in dynamic environments. 
% This approach provides new insights into harnessing event data for fast and efficient object detection, outperforming traditional methods in time-sensitive scenarios. 
SAFE~\cite{li2023semantic} introduces an innovative framework that integrates semantic labels, RGB frames, and event streams. By leveraging a large pre-trained vision-language model, this approach addresses the semantic gap and overcomes the limitations associated with small-scale backbone networks in traditional methods. However, event cameras have not been widely explored in scene text recognition. Leveraging their advantages, such as high temporal resolution and low power consumption, we propose a method to use event data for text recognition, aiming to improve performance in dynamic and challenging environments where traditional methods face difficulties.

\section{Our Proposed Approach} \label{sec::method}

\begin{figure*}
\centering
\includegraphics[width=0.9\linewidth]{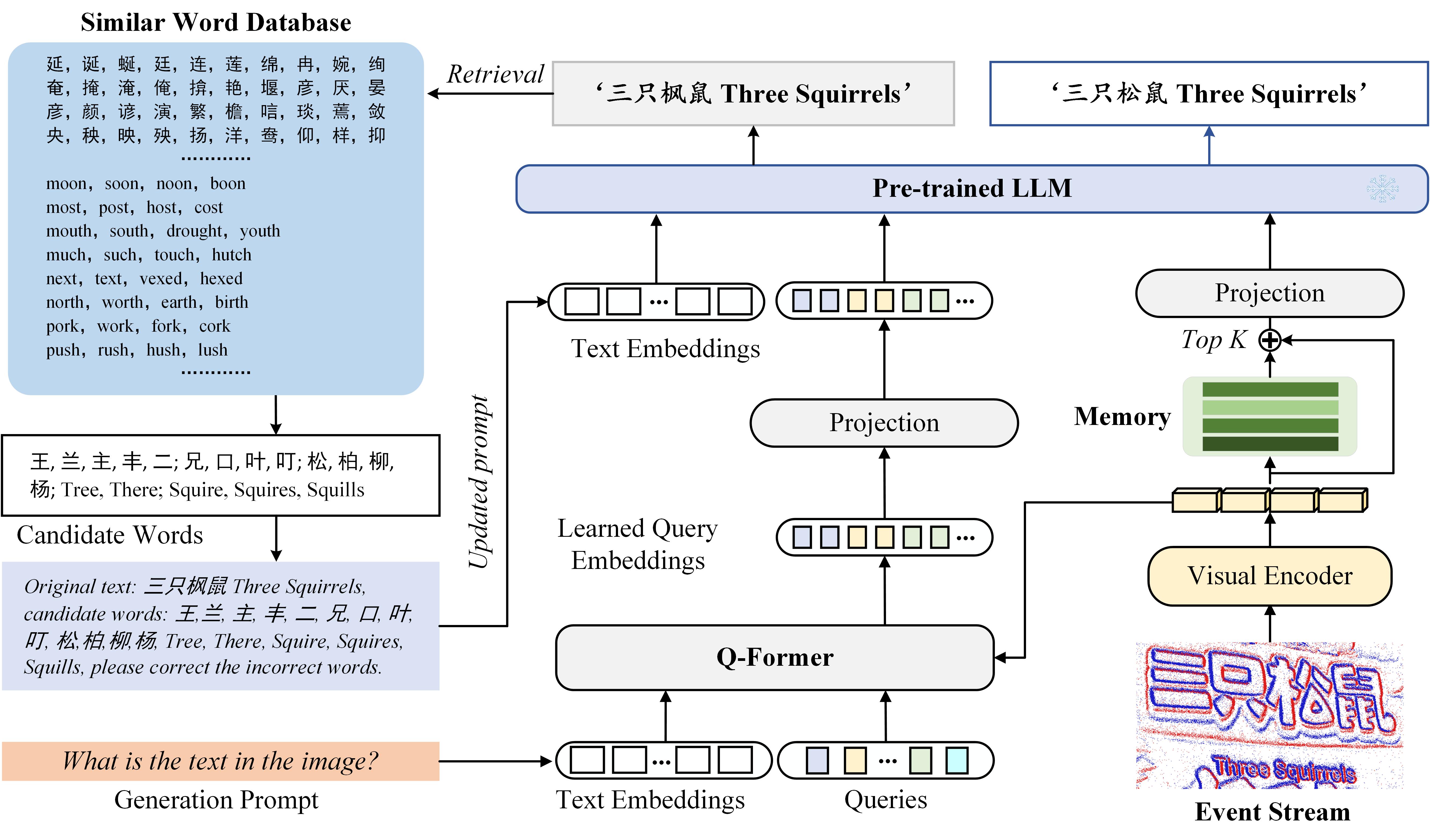}
\caption{An overview of our proposed large language model based event stream scene text recognition framework, termed SimC-ESTR.
Given the event streams, we first stack them into a single event frame and use a visual encoder to extract feature representations. These features are passed through a Q-former module to align vision tokens with a pre-trained large language model (LLM), which then generates text. To further enhance the features, we introduce a memory mechanism that leverages contextual samples for better representation. We also address the issue of LLMs occasionally producing incorrect but visually similar Chinese characters by designing a correction module specifically for such cases. More details of these modules will be described in Section~\ref{sec::network}. }  
\label{fig:framework}
\end{figure*}

In this section, we will first give an overview of our framework. Then, we focus on the detailed network architectures, including the Visual Encoder, Memory Mechanism, Glyph Error Correction Module, and Pre-trained Large Language Model. After that, we describe the loss function used for the optimization of our framework.

\subsection{Overview}  
Considering that existing scene text recognition algorithms are mostly based on RGB frames, in order to better adapt to these models, we also adopt the approach of stacking event streams into event frames for experimentation in this paper. Specifically, we first stack them into a single event frame by following the method used in EventVOT~\cite{wang2024eventvot}. Then, we adopt a visual encoder to embed the input into feature representations. The features are fed into the Q-former to align the vision tokens and large language model and the output tokens are fed into a pre-trained large language model for text generation. Meanwhile, we propose to utilize the memory mechanism to augment the features further. This is mainly because text symbols have similarities, and the text symbols in the contextual samples can also provide a reference for the representation of the current symbol. We have observed that large language models sometimes output incorrect Chinese characters, but these characters are indeed very similar to the correct ones, i.e., they are visually similar characters. Therefore, we have designed a set of visually similar character correction modules to help large language models produce more accurate text recognition results. More details will be introduced in the subsequent subsections.

\subsection{Input Representation}  
The event stream $\mathcal{E} = \{\textit{e}_1, \textit{e}_2, ..., \textit{e}_N\}$ can be seen as a spatial-temporal flow similar to the point cloud~\cite{jiang2023masked}, each event point $\textit{e}_i$ can be denoted as [$x, y, t, p$]. Here, $N$ is the number of points in a single event stream. $(x, y)$ is the spatial coordinates, $t$ is the timestamp, $p \in \{1, -1\}$ denotes the polarity (positive/negative) of the event point. As mentioned in the previous section, we stack the event streams into event frames $\mathcal{I} \in \mathbb{R}^{T \times C \times H \times W} = \{\textit{I}_1, \textit{I}_2, ..., \textit{I}_T\}$ to better adapt existing STR models for benchmark comparison, where $T$ is the number of stacked event frames. Stacking event streams into frames akin to RGB frames offers key advantages such as compatibility with existing algorithms and toolchains, explicit modeling of temporal information, improved spatial feature extraction efficiency, mitigation of data sparsity and noise issues, and support for intuitive visualization, making it a practical and efficient solution well-suited for scenarios requiring rapid development and deployment. 
Due to the utilization of a large language model, in this work, we also take a generation prompt $\mathcal{P}$ as the input, i.e., ``\textit{What is the text in the image?}''. The LlamaTokenizer~\cite{touvron2023llama} is used to get the text embeddings $F_l$ for further processing.

\subsection{Network Architecture} \label{sec::network}

The key modules of our proposed SimC-ESTR framework are the Visual Encoder, Memory Module, Pre-trained LLM, and Glyph Error Correction Module, as shown in Fig.~\ref{fig:framework}.

\noindent $\bullet$ \textbf{Visual and Prompt Encoder.~} 
Given the event frames $\mathcal{I}$, we adopt the pre-trained EVA-CLIP~\cite{sun2023eva} model (ViT-G/14) as the visual encoder for feature extraction. Specifically, it processes input images by dividing them into fixed-size patches (14$\times$14 pixels), which are flattened into tokens for long-range spatial representation. The key operator is the multi-head self-attention mechanism which focuses on discriminative features and the output visual feature can be denoted as $F_v$. It also outputs a global token [CLS] for representation of the whole input frame. These visual features are further refined by the Q-Former and projected into the LLM for final text recognition.

For the generation prompt $\mathcal{P}$, we propose the text encoder to guide the model in understanding and recognizing the visual content from the event-based image. The textual prompt is tokenized and transformed into a text embedding. Here, the prompt is tokenized into token IDs, which are then passed through different tokenizers depending on the part of the model. The first tokenizer processes the prompt for use by the Q-Former, which integrates visual and textual information. The second tokenizer prepares the prompt for the LLM, which generates the final text predictions. The tokenized prompt \( F_l \) is used by both the Q-Former and the LLM, where the embedding representations may differ due to optimization for their respective components. The tokenization process ensures that the prompt is properly formatted for integration with the model components and their attention mechanisms, enabling smooth interaction between textual and visual features. By using the tokenized prompt, we ensure the efficient integration of textual context with event-based visual features, optimizing the model's performance in generating accurate scene text recognition results.

\noindent $\bullet$ \textbf{Memory Module.~}
It is designed to enhance the model's ability to capture long-term dependencies by leveraging a pattern-based memory mechanism. It consists of a set of learnable memory patterns that are used to improve the input features through mapping, similarity matching, and feature enhancement. More in detail, the input visual features, which have dimensions \( B \times L \times D \) (where \( B \) represents the batch size, \( L \) is the sequence length, and \( D \) denotes the feature dimension), are reshaped and passed through a linear layer. This linear layer projects the features into a lower-dimensional space, specifically into a 128-dimensional space, which corresponds to the predefined pattern dimension. After that, the module computes the cosine similarity between the projected input features and a set of stored memory patterns. These patterns, which are initialized randomly and are learnable, capture key visual representations learned during training. The module selects the top-$K$ most similar patterns from the memory, and these patterns are transformed back into the original feature space using another linear layer. This process generates a set of enhanced features.

The final output is obtained by adding the weighted average of the top-$K$ most relevant patterns to the original input features. It allows the model to improve its predictive capabilities, particularly in the case of incomplete or noisy input data. The memory mechanism enables the model to store and recall important visual representations over time, thereby improving performance in long-term visual perception tasks such as text recognition in complex environments.

\noindent $\bullet$ \textbf{Pre-trained LLM.~} 
In this work, we adopt the LLaMA-based LLM fine-tuned with supervised instruction data, i.e., Vicuna-7B~\cite{touvron2023llama}, for scene text recognition. 
Note that, the LLM Vicuna-7B is frozen during the training and testing phase. The LLM receives a multi-modal input that combines three components, including prompt embedding $F_l$, learnable query embeddings from the Q-Former, and visual features extracted from the image. These components are projected into a unified feature space and concatenated as follows: 
\begin{equation}
\text{Output} = \text{LLM}([F_l, \text{Proj}(Q), \text{Proj}(F_v)])
\end{equation}
where \( \text{Proj}(Q) \) is the projected output of the learnable query embeddings, and \( \text{Proj}(F_v) \) denotes the projected visual features. 
This fusion enables the LLM to effectively integrate prompt information with visual context, improving text recognition accuracy in complex scenes.

\noindent $\bullet$ \textbf{Glyph Error Correction Module.~} 
Event-based images excel in capturing dynamic scenes and performing well in low-light conditions. However, they often come with inherent limitations that can lead to recognition errors. Due to their sparse, noisy, and incomplete nature, stemming from the fact that they only capture changes in the scene, text characters may appear fragmented, distorted, or ambiguous. These issues can significantly increase the risk of misrecognition during the initial text prediction phase.

To address this issue, we propose the Glyph Error Correction Module, designed to enhance recognition accuracy through glyph-based corrections. This module operates in two key stages: first, constructing a visually similar glyph database, and second, correcting ambiguous characters based on this database, which will be introduced in the subsequent paragraphs. 

\textbf{1) Similar Glyph Database Construction:}  
We construct the similar glyph database by initially collecting visually similar character pairs from publicly available online resources. The construction process involves: 

\textit{- Online Collection:} Gathering similar character sets and word lists from various linguistic resources and databases available on the internet, covering both Chinese and English.  

\textit{- Manual Refinement:} Carefully reviewing the collected data to add, remove, or adjust character pairs based on their visual resemblance. This step ensures the inclusion of task-relevant glyphs.  

\textit{- Task-Specific Adjustment:} Modifying the database according to the recognition errors observed in preliminary experiments. This helps optimize the database for event-based scene text recognition scenarios, enhancing the correction module’s performance.

\textbf{2) Glyph-Based Error Correction:}
After generating the initial text prediction, the module performs a character-wise analysis to identify potentially erroneous glyphs. For each character:
\begin{CJK*}{UTF8}{gbsn}
\begin{itemize}
    \item \textbf{Visually Similar Character Retrieval:} 
    We query the similar glyph database to retrieve a list of visually similar candidates. 
    Consider a glyph database where similar characters are grouped based on visual resemblance. 
    For instance, for the character '苍', the following visually similar candidates could be retrieved, i.e., \{\text{沧}, \text{抢}, \text{枪}\}. 
    Similarly, for the character '吹', potential similar characters could include \{\text{炊}, \text{饮}, \text{欢}\}. 
    For English characters, common visually similar candidates might include: 
    \text{candidates for "cap"}: \{\text{map}, \text{nap}, \text{lap}\}; 
    \text{candidates for "deed"}: \{\text{need}, \text{seed}, \text{reed}\}. 
    
    \item \textbf{Contextual Validation:} 
    To avoid introducing new errors, the retrieved candidates are validated using contextual information from the surrounding text, ensuring semantic coherence. 
    \item \textbf{Prompt Update:} The corrected characters are used to update the initial prompt, forming the refined prompt $\tilde{\mathcal{P}}$, which is re-encoded as  $\tilde{F_l}$. 
\end{itemize}
\end{CJK*}
The final LLM input, incorporating glyph corrections, is expressed as: 
\begin{equation} 
\tilde{\text{Output}} = \text{LLM}([\tilde{F_l}, \text{Proj}(Q), \text{Proj}(F_v)])
\end{equation}
Here, $\tilde{F_l}$ represents the corrected text embeddings, $\text{Proj}(Q)$ is the projected query embeddings from the Q-Former, and $\text{Proj}(F_v)$ denotes the visual features. This comprehensive input, integrating visual cues, contextual information, and glyph-based corrections, significantly improves scene text recognition accuracy. For an in-depth analysis of prompt variations and their impact on error correction, please refer to Section~\ref{prompt}.

\begin{table*}
\centering
\small 
\caption{Comparison of Datasets for Scene Text Recognition. TS represents the dataset structure, DT represents the data type, MS and M-ILL denote multi-scenario and multi-illumination, respectively. TO represents text orientation, H and V indicate whether horizontal and vertical text are present.}
\label{tab:datasets}
\begin{tabular}{lcl|ccc|c|c|c|c|cc}
\toprule
\multirow{2}{*}{\textbf{Dataset}} & \multirow{2}{*}{\textbf{Conf.}} & \multirow{2}{*}{\textbf{Year}} & \multicolumn{3}{c|}{\textbf{\# of word boxes}} & \multirow{2}{*}{\textbf{TS}}& \multirow{2}{*}{\textbf{DT}}&\multirow{2}{*}{\textbf{MS}}&\multirow{2}{*}{\textbf{M-ILL}}& \multicolumn{2}{c}{\textbf{TO}}\\
\cmidrule(lr){4-6} \cmidrule(lr){11-12}
& & & \textbf{Train} & \textbf{Val} & \textbf{Test}& & & & & \textbf{H}&\textbf{V}\\
\midrule
\multicolumn{12}{l}{\textbf{Synthetic datasets}} \\
MJ~\cite{jaderberg2014MJ} & NIPSW & 2014 & 7,224,586 & 802,731& 891,924 & Regular& RGB& & & \ding{51}&\\
ST~\cite{gupta2016ST} & CVPR & 2016 & 6,975,301 & - & -  & Regular& RGB& & & \ding{51}&\\
\midrule
\multicolumn{12}{l}{\textbf{Real datasets}}  \\
SVT~\cite{wang2011SVT} & ICCV & 2011 & 257 & - & 647  & Regular& RGB& & & \ding{51}&\\
IIIT5k~\cite{mishra2012III5k} & BMVC & 2012 & 2,000 & - & 3,000  & Regular& RGB& \ding{51}& & \ding{51}&\\
IC13~\cite{karatzas2013icdar} & ICDAR & 2013 & 848 & - & 1,015  & Regular& RGB& & & \ding{51}&\\
SVTP~\cite{phan2013svtp} & ICCV & 2013& - & - & 645  & Irregular& RGB& & & \ding{51}&\\
CUTE~\cite{risnumawan2014cute} & ESWA & 2014 & - & - & 288  & Irregular& RGB& \ding{51}& & \ding{51}&\\
IC15~\cite{karatzas2015icdar} & ICDAR & 2015 & 4,468 & - & 2,077  & Irregular& RGB& \ding{51}& & \ding{51}&\\
COCO~\cite{veit2016coco} & arXiv & 2016 & 59,820 & 13,415 & 9,825  & Irregular& RGB& \ding{51}& & \ding{51}&\\
RCTW17~\cite{shi2017RCYW17} & ICDAR & 2017 & 8,034 & - & 10,509  & Regular& RGB& \ding{51}& & \ding{51}&\\
Uber~\cite{zhang2017uber} & CVPRW & 2017 & 91,378 & 36,136 & 80,914  & Irregular& RGB& \ding{51}& & \ding{51}&\\
ArT~\cite{chng2019ArT} & ICDAR & 2019 & 32,349 & - & 35,149  & Irregular& RGB& \ding{51}& & \ding{51}&\\
ReCTS~\cite{zhang2019ReCTS} & ICDAR & 2019 & 20,000 & - & 2,592  & Irregular& RGB& & & \ding{51}&\ding{51}\\
LSVT~\cite{sun2019LSVT} & ICDAR & 2019 & 43,244 & - & -  & Irregular& RGB& & & \ding{51}&\ding{51}\\
MLT19~\cite{nayef2019MLT19} & ICDAR & 2019 & 56,937 & - & 9,896  & Irregular& RGB& & & \ding{51}&\\
TextOCR~\cite{singh2021textocr} & ECCV & 2020 & 714,770 & 107,722 & -  & Irregular& RGB& & & \ding{51}&\\
WordArt~\cite{xie2022CornerTransformer}& ECCV& 2022& 4805& -& 1511& Irregular& RGB& \ding{51}& & \ding{51}&\ding{51}\\
Union14M~\cite{MAERec}& ICCV & 2023 & - & - & 403,379  & Irregular& RGB& \ding{51}& & \ding{51}&\\
\hline 
EventSTR (Ours)& -& 2025 & 6949& 993&1986  & Irregular& Event& \ding{51}& \ding{51}& \ding{51}&\ding{51}\\
\bottomrule
\end{tabular}
\end{table*}

\section{EventSTR Benchmark Dataset} \label{sec::EventSTR-benchmark}
In this paper, we introduce a new event-based scene text recognition dataset, termed \textbf{EventSTR}. The following paragraphs provide a detailed description of the data collection and annotation process, statistical analysis, and the benchmark protocols for visual trackers.

\begin{figure*}[!htp]
\centering
\includegraphics[width=\linewidth]{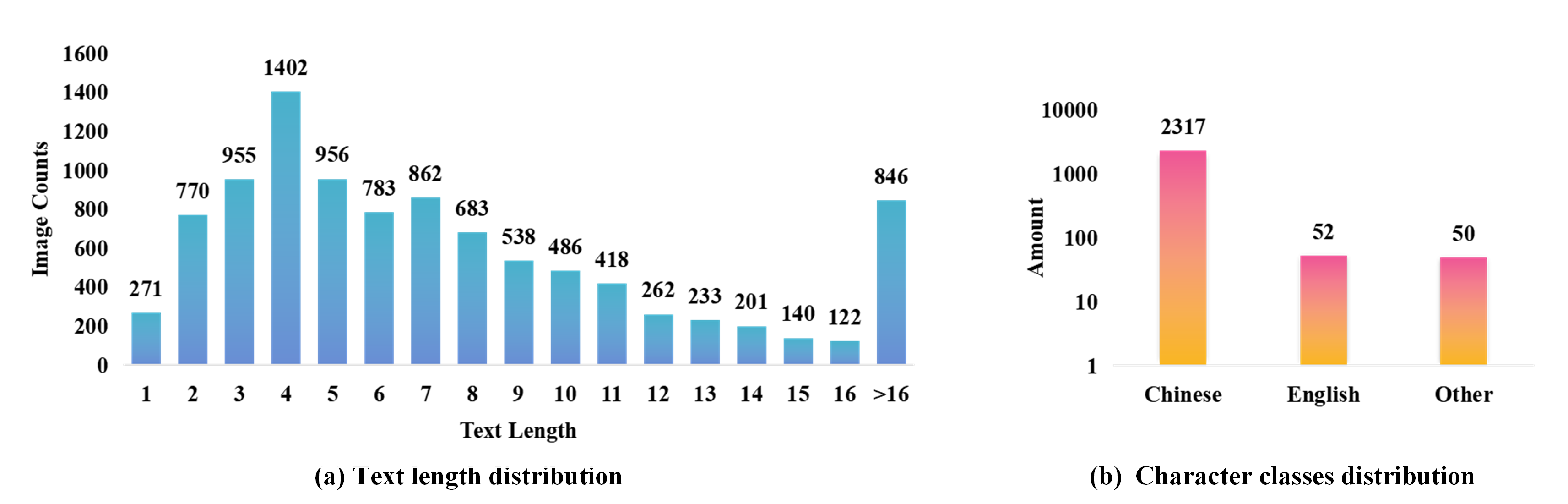}
\caption{Statistical analysis for the EventSTR dataset. (a) The number of images with
different text lengths. (b) Distribution of the number of characters.}
\label{fig:dataset_distribution}
\end{figure*}

\subsection{Protocols}   
We aim to provide a new direction for scene text recognition using event-based data. The EventSTR benchmark dataset was constructed adhering to the following protocols:
\textit{1). Lighting Conditions:} The dataset was captured under challenging low-light conditions, where traditional image-based methods would struggle. However, thanks to the high sensitivity of the event camera, text remains clearly visible even in dark scenes with low light intensity.
\textit{2). Motion Variability:} The dataset includes images captured at varying motion speeds, resulting in scenarios where text may appear blurred or distorted due to motion, adding complexity to text recognition tasks.
\textit{3). Occlusion:} The dataset features images with varying levels of occlusion, where portions of the text may be obstructed, making the recognition task more difficult.
\textit{4). Scene Categories:} A wide range of scene categories is included, such as posters, books, commodities, billboards, and license plates, providing diverse real-world scenarios.
\textit{5). High Resolution:} The dataset is captured using the Prophesee Evaluation Kit 4 HD (EVK4) event camera~\footnote{\url{https://www.prophesee.ai/event-based-sensors/}}, with a resolution of 1280$\times$720, ensuring high-quality image details.
\textit{6). Text Orientation:} The dataset contains both horizontal and vertical text orientations, in contrast to most other datasets that typically feature only horizontal, single-line text. This diversity in text orientation introduces additional challenges in text recognition tasks.

\begin{figure}
\centering
\includegraphics[width=\linewidth]{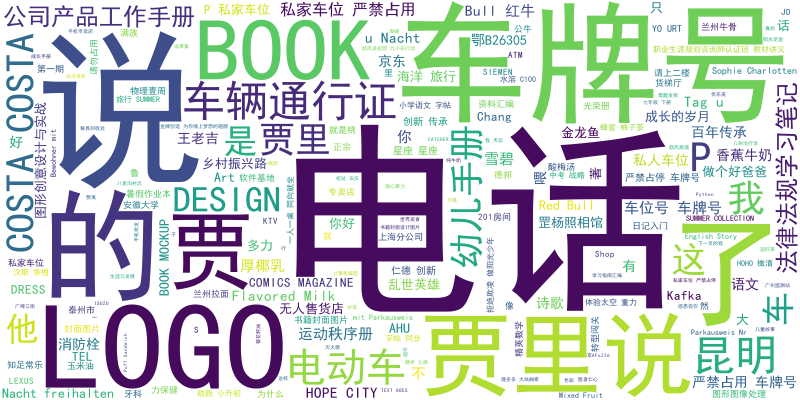}
\caption{The word cloud visually represents the frequency distribution of words in the EventSTR dataset labels. Words that appear more frequently are displayed larger and more prominently, whereas smaller words correspond to those with lower occurrence rates.}
\label{fig:wordcloud}
\end{figure}

\subsection{Data Collection and Annotation} 
EventSTR dataset was collected using the Prophesee HD event camera, featuring a resolution of 1280 $\times$ 720. We adhered to a specific protocol during the data collection process. We followed the following principles during the annotation process: 
\textit{(1)} All characters occurring in the scene must be labeled; 
\textit{(2)} The labeling must exactly match the text as it appears in the scene; 
\textit{(3)} No annotations are made for scenes that are excessively dark or have motion blur, as these conditions hinder text recognition.

\subsection{Statistical Analysis} 
From a statistical perspective, our dataset consists of 9,928 video sequences and encompasses a total of 2,300 character classes. During the data processing stage, each video sequence is converted into 19 event frames, with the first frame selected as the final representation of the dataset. The dataset is then divided into training, validation, and test sets in a ratio of 7:1:2, resulting in 6,949 images for training, 993 images for validation, and 1,986 images for testing. We also analyzed the number of images corresponding to different text lengths, as shown in Fig.~\ref{fig:dataset_distribution} (a). The dataset contains a total of 2,317 Chinese characters, while the English characters consist only of 26 uppercase letters and 26 lowercase letters. Additionally, there are 50 other characters, as illustrated in Fig.~\ref{fig:dataset_distribution} (b). We also provide a word cloud visualization to illustrate the frequency distribution of characters in Fig.~\ref{fig:wordcloud}.

\begin{figure*}[!htp]
    \centering
    \includegraphics[width=\linewidth]{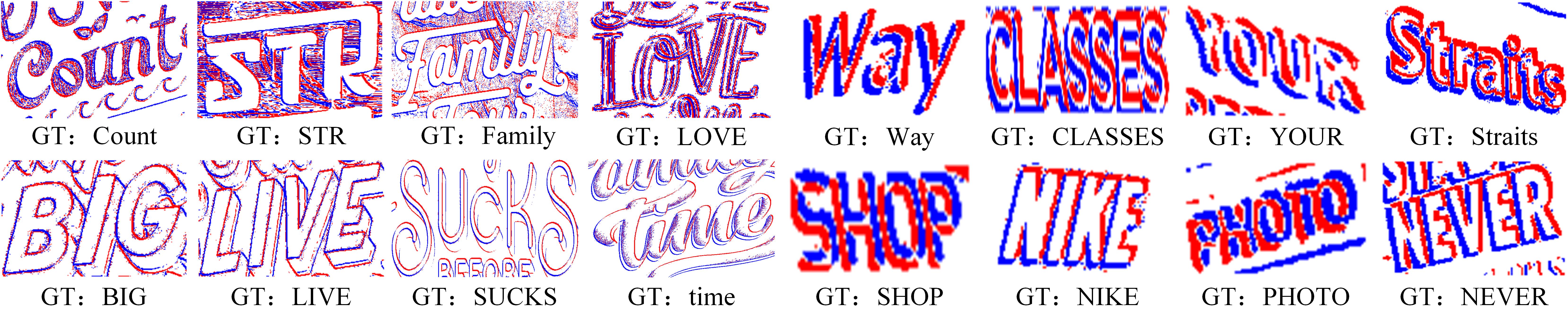}
    \caption{Illustration of representative samples of the synthetic WordArt* and IC15* dataset.}
    \label{fig:synthetic_OCR_dataset_VIS}
\end{figure*}

\subsection{Benchmark Baseline} 
To build a comprehensive benchmark dataset for event-based scene text recognition, we include the following text recognition models: LISTER~\cite{cheng2023lister}, CCD~\cite{Guan2023CCD}, SIGA~\cite{guan2023SIGA}, CDistNet~\cite{zheng2024cdistnet}, DiG~\cite{yang2022DiG}, PARSeq~\cite{bautista2022PARseq}, MGP-STR~\cite{wang2022mgpstr}, and OCR2.0~\cite{wei2024OCR2.0}. These models, initially trained on two large text recognition datasets (MJ~\cite{jaderberg2014MJ} and ST~\cite{gupta2016ST}), are fine-tuned on the training subsets of three datasets: EventSTR, WordArt~\cite{xie2022CornerTransformer}, and IC15~\cite{karatzas2015icdar}, and evaluated on their respective test subsets. For OCR2.0 specifically, we load its pre-trained weights and then fine-tune it on the training subset of each dataset. We believe that these fine-tuned scene text recognition models will be essential for future performance evaluations.

\section{Experiments} \label{sec::Exp}

\subsection{Dataset and Evaluation Metric}  
For the datasets, we evaluate our model alongside other state-of-the-art methods on three main datasets: \textbf{WordArt*}~\footnote{\url{https://opendatalab.com/OpenDataLab/WordArt}}, \textbf{IC15*}~\footnote{\url{https://aistudio.baidu.com/datasetdetail/96799}}, and our newly introduced \textbf{EventSTR}. Below is a brief introduction to each of these datasets. Details of the other datasets are provided in Table~\ref{tab:datasets}. 

$\bullet$ \textbf{WordArt* Dataset:} As shown in Fig.~\ref{fig:synthetic_OCR_dataset_VIS}, this dataset is derived from the original RGB-format WordArt~\cite{xie2022CornerTransformer} dataset and simulated into event-based images using event camera simulator (ESIM)~\cite{rebecq2018esim}. It is split into a training set of 4,805 images and a validation set of 1,511 images. The dataset contains artistic text images, including posters, greeting cards, covers, billboards, handwritten texts, and more. These images feature a variety of artistic text styles. 

$\bullet$  \textbf{IC15* Dataset:} This dataset is created by transforming the original RGB-format IC15~\cite{karatzas2015icdar} dataset into event-based images. IC15* is a natural scene text dataset, consisting of 4,468 training images and 2,077 testing images.

For the evaluation metrics, we use BLEU-1, BLEU-2, BLEU-3, and BLEU-4 scores to assess performance on the EventSTR dataset, which involves multi-text scenarios. BLEU scores are calculated by segmenting characters for Chinese text and by words (case-insensitive) for English text. For the WordArt* and IC15* datasets, we employ the word-level recognition accuracy.

\subsection{Implementation Details}  
We use the pre-trained weights from BLIVA~\cite{hu2024bliva}, followed by fine-tuning on our dataset. The AdamW~\cite{loshchilov2017AdamW} optimizer was employed, with $\beta_1 = 0.9$, $\beta_2 = 0.999$, and a weight decay of $0.05$. Additionally, we apply a linear warm-up for the learning rate over the first 1,000 steps, gradually increasing it from $10^{-8}$ to $10^{-5}$, followed by a cosine decay to a minimum learning rate of $0$. All experiments are conducted on an Nvidia A800 GPU. More details can be found in our source on GitHub.

\begin{table*}
\centering
\small 
\caption{Comparison of BLEU scores with SOTA methods on the EventSTR dataset.}
\label{tab:bleu}
\resizebox{\textwidth}{!}{ 
\begin{tabular}{l|l|c|cccccc}
\hline 
\textbf{Algorithm}  & \textbf{Publish}  & \textbf{Backbone}  & \textbf{BLEU-1} & \textbf{BLEU-2} & \textbf{BLEU-3} & \textbf{BLEU-4}  &\textbf{Params(M)}& \textbf{Code}   \\ \hline
CCD~\cite{Guan2023CCD}  & ICCV 2023  & ViT  &  0.365 & 0.254 & 0.172 & 0.145 &52.0& \href{https://github.com/TongkunGuan/CCD}{URL}\\
SIGA~\cite{guan2023SIGA}  & CVPR 2023  & ResNet&  0.434& 0.393& 0.346& 0.307 &40.4& \href{https://github.com/TongkunGuan/SIGA}{URL}\\
CDistNet~\cite{zheng2024cdistnet}  & IJCV 2023  & ResNet+Transformer&  0.333 & 0.242 & 0.157 & 0.135 &65.5& \href{https://github.com/simplify23/CDistNet?tab=readme-ov-file}{URL}  \\
PARSeq~\cite{bautista2022PARseq}  & ECCV 2022  & ViT  &  0.450& 0.357& 0.281& 0.224 &23.4& \href{https://github.com/baudm/parseq}{URL}\\
MGP-STR~\cite{wang2022mgpstr}& ECCV 2022& Transformer& 0.427& 0.339& 0.278& 0.232 &148.0&\href{https://github.com/AlibabaResearch/AdvancedLiterateMachinery/tree/main/OCR/MGP-STR}{URL}\\
GOT-OCR2.0~\cite{wei2024OCR2.0}& arXiv 2024 &ViT & 0.426 & 0.390 & 0.358 & 0.332  &580.0&\href{https://github.com/Ucas-HaoranWei/GOT-OCR2.0}{URL}\\  
% \cline{1-9}
BLIVA~\cite{hu2024bliva}&  AAAI 2024 & ViT  &0.584 & 0.528 & 0.450 & 0.386 &7531.3& \href{https://github.com/mlpc-ucsd/BLIVA}{URL}  \\
\hline
SimC-ESTR (Ours)&-& ViT  & \textbf{0.638}& \textbf{0.583} & \textbf{0.500} & \textbf{0.430} &7531.3& \href{https://github.com/Event-AHU/EventSTR}{URL} \\ 
\hline
\end{tabular}} 
\end{table*}

\begin{table*}
\centering 
\small 
\caption{The accuracy comparisons with SOTA methods on WordArt* and IC15*.}
\label{tab:acc}
\begin{tabular}{l|l|c|c|c|c|c}
\hline 
\multirow{2}{*}{\textbf{Algorithm}}  & \multirow{2}{*}{\textbf{Publish}}  & \multirow{2}{*}{\textbf{Backbone}}  & \multicolumn{2}{c|}{\textbf{Accuracy}} &  \multirow{2}{*}{\textbf{Params(M)}}&\multirow{2}{*}{\textbf{Code}} \\
\cline{4-5}  
&  &  & \textbf{WordArt*}& \textbf{IC15*}&   &\\
\hline 
LISTER~\cite{cheng2023lister}& ICCV 2023& CNN& 55.3& 69.0 &  49.9&\href{https://github.com/AlibabaResearch/AdvancedLiterateMachinery/tree/main/OCR/LISTER}{URL}\\
CCD~\cite{Guan2023CCD}  & ICCV 2023  & ViT  &  62.1&  55.4&  52.0&\href{https://github.com/TongkunGuan/CCD}{URL} \\
SIGA~\cite{guan2023SIGA}  & CVPR 2023  & ResNet&  69.0&  66.2&  40.4&\href{https://github.com/TongkunGuan/SIGA}{URL} \\
CDistNet~\cite{zheng2024cdistnet}  & IJCV 2023  & ResNet+Transformer  &  66.6&  62.3&  65.5&\href{https://github.com/simplify23/CDistNet?tab=readme-ov-file}{URL} \\
DiG~\cite{yang2022DiG}  & ACM MM 2022  & ViT  &  62.7&  53.2&  52.0&\href{https://github.com/ayumiymk/DiG}{URL} \\
PARSeq~\cite{bautista2022PARseq}  & ECCV 2022  & ViT  &  75.0&  72.7&  23.4&\href{https://github.com/baudm/parseq}{URL} \\
MGP-STR~\cite{wang2022mgpstr}& ECCV 2022& Transformer& 69.6& 67.5& 148.0&\href{https://github.com/AlibabaResearch/AdvancedLiterateMachinery/tree/main/OCR/MGP-STR}{URL}\\
% \cline{1-7} 
BLIVA~\cite{hu2024bliva} &  AAAI 2024& ViT  &  56.7&  51.3&  7531.3&\href{https://github.com/mlpc-ucsd/BLIVA}{URL} \\
\hline 
SimC-ESTR (Ours) &  -& ViT  &  65.1&  56.8&  7531.3&\href{https://github.com/Event-AHU/EventSTR}{URL} \\
\hline 
\end{tabular}
\end{table*}

\subsection{Comparison on Public Benchmark Datasets} 

\noindent $\bullet$ \textbf{Results on EventSTR Dataset.~} 
Table~\ref{tab:bleu} compares the BLEU scores of our method with several SOTA algorithms on the EventSTR dataset. Our method achieves significant improvements across all BLEU metrics. Specifically, it outperforms the baseline and other SOTA methods in BLEU-1, BLEU-2, BLEU-3, and BLEU-4, with BLEU scores of 0.629, 0.570, 0.486, and 0.417, respectively. These results demonstrate the effectiveness of our approach in capturing and generating accurate text representations. Overall, the results highlight the strong performance of our method in scene text recognition tasks, particularly in handling complex and diverse text appearances, as found in the EventSTR dataset.

\noindent $\bullet$ \textbf{Results on WordArt* and IC15* Datasets.~} 
As shown in Table~\ref{tab:acc}, our method, which was pre-trained on Visual Question Answering (VQA) data, does not achieve optimal accuracy on both WordArt* and IC15* datasets compared to methods trained on large-scale text recognition datasets (such as MJ and ST). While VQA pre-training supports the model’s understanding of visual-textual relationships, it may not be ideally suited for text recognition tasks, particularly in complex or noisy backgrounds, which are better handled by models trained specifically on OCR data. 

Moreover, the synthetic datasets (WordArt* and IC15*) used for fine-tuning our model are relatively low in resolution and lack the diversity and complexity typically present in larger OCR datasets. This limited dataset quality and scope likely contributed to our model’s performance not reaching the level of methods such as LISTER, CCD, and PARSeq, which benefit from training on extensive, high-quality OCR datasets.

% \textcolor{red}{--- Xiao Wang} 

\subsection{Component Analysis} 

As shown in Table~\ref{tab:component analysis}, two key modules are separately validated on the proposed EventSTR dataset, i.e., Glyph Error Correction Module (GECM) and Memory Module (MM). It is easy to find that the baseline model (line \#01), which excludes both GECM and MM, achieves a BLEU-1 score of 0.584. Adding GECM alone (line \#02) improves the BLEU-1 score to 0.629, reflecting an absolute improvement of 0.045. This significant gain highlights the ability of GECM to effectively address visually similar glyph errors, thereby enhancing recognition accuracy. When incorporating MM alone (line \#03), the BLEU-1 score increases to 0.608, showing an improvement of 0.024 over the baseline. This result demonstrates the role of MM in enriching feature representations by utilizing stored patterns. Combining both GECM and MM (line \#04) achieves the highest BLEU-1 score of 0.638, which represents an absolute improvement of 0.054 over the baseline. These results emphasize the complementary nature of GECM and MM, as their combination consistently improves performance across BLEU-1 as well as higher-order BLEU metrics.

\newcommand{\yes}{\textcolor{SeaGreen4}{\ding{51}}}
\newcommand{\no}{\textcolor{DarkRed}{\ding{55}}}

\begin{table}[!htp]
\caption{Component analysis of the key modules in our framework on EventSTR dataset.}
\label{tab:component analysis}
\centering
\resizebox{0.49\textwidth}{!}{ 
\begin{tabular}{c|cc|cccc}
\hline
\textbf{No.} & \textbf{GECM} & \textbf{MM} & \textbf{BLEU-1} & \textbf{BLEU-2} & \textbf{BLEU-3} & \textbf{BLEU-4} \\ \hline
\#01 & \no  & \no  & 0.584 & 0.528 & 0.450 & 0.386 \\
\#02 & \yes & \no  & 0.629 & 0.570 & 0.486 & 0.417 \\
\#03 & \no  & \yes &  0.608 &  0.548 &  0.466 &  0.398 \\
\#04 & \yes & \yes &  \textbf{0.638}&  \textbf{0.583} &  \textbf{0.500} &  \textbf{0.430} \\ 
\hline
\end{tabular}} 
\end{table}

\subsection{Ablation Study}

\noindent $\bullet$ \textbf{Impact of Top-K Selection in Memory Module.~}  
The Top-K parameter in the Memory Module determines the number of most similar patterns retrieved from the memory pool for feature enhancement. As shown in Table~\ref{tab:top-k}, varying the value of $K$ directly impacts the BLEU scores, reflecting the module's ability to effectively recall and utilize learned patterns. When $K=3$, the BLEU scores are relatively low due to insufficient diversity in the retrieved patterns, which limits the module's capacity to enrich the input features. Increasing $K$ to 32 and 64 leads to significant improvements across all BLEU metrics, as a larger number of patterns provides more comprehensive contextual information, allowing for better feature refinement. However, when $K$ is further increased to 128, the BLEU scores slightly drop, suggesting that including too many patterns may introduce noise or redundant information, diluting the effectiveness of the memory mechanism. This analysis demonstrates that selecting an appropriate $K$ value is crucial for balancing the diversity of retrieved patterns and avoiding potential overfitting or information redundancy. In this case, $K=64$ achieves the best overall performance, providing an optimal trade-off between pattern diversity and feature enhancement.

\begin{table}
\caption{BLEU Score Comparison Across Different Top-K Values in Memory Module.}
\label{tab:top-k}
\centering
\begin{tabular}{c|cccc}
\hline
\textbf{K}&  \textbf{BLEU-1} &  \textbf{BLEU-2} &  \textbf{BLEU-3} & \textbf{BLEU-4} \\ 
\hline
3&  0.606 &  0.546 &  0.464 & 0.400 \\
32&  0.633&  0.574 &  0.491 & 0.423 \\
64&  \textbf{0.638}&  \textbf{0.583} &  \textbf{0.500} & \textbf{0.430} \\
128&0.624 &0.563 &0.480 &0.411  \\ 
\hline
\end{tabular}
\end{table}

\noindent $\bullet$ \textbf{Analysis of Different Prompts for Error Correction.~} \label{prompt} 
\begin{CJK*}{UTF8}{gbsn}
In this experiment, we explore the impact of different prompt phrasings on the model's text correction performance. The goal is to assess how varying prompt styles influence the model's accuracy, consistency, and effectiveness in error correction across different contexts. We use three distinct prompt formulations as follows, with "三只枫鼠 Three Squirrels" as an example: 

\textbf{Prompt 1:} \textit{The following text may contain errors: 三只枫鼠 Three Squirrels. Possible replacements include: 王, 兰, 主, 丰, 二, 兄, 口, 叶, 叮, 松, 柏, 柳, 杨, Tree, There, Squire, Squires, Squills. Please make corrections.
}

\textit{Objective:} Explicitly informs the model that the text may contain errors and provides possible replacements, indicating that corrections are required.

\textbf{Prompt 2:} \textit{Correct the text: '三只枫鼠 Three Squirrels'. Use these candidates for guidance: 王, 兰, 主, 丰, 二, 兄, 口, 叶, 叮, 松, 柏, 柳, 杨, Tree, There, Squire, Squires, Squills.}

\textit{Objective:} Directly instructs the model to correct the text, emphasizing the use of candidate words as a guide.

\textbf{Prompt 3:} \textit{Original text: 三只枫鼠 Three Squirrels, candidate words: 王, 兰, 主, 丰, 二, 兄, 口, 叶, 叮, 松, 柏, 柳, 杨, Tree, There, Squire, Squires, Squills, please correct the incorrect words.}

\textit{Objective:} Provides the original text and candidate words in a conversational style, suggesting correction without explicitly enforcing it.
\end{CJK*}

\begin{table}
\centering
\small 
\caption{BLEU Scores of Different Prompts.} 
\label{tab:prompt}
\begin{tabular}{c|cccc}
\hline
\textbf{Prompt}&  \textbf{BLEU-1} &  \textbf{BLEU-2}&  \textbf{BLEU-3}& \textbf{BLEU-4}\\ \hline
\#1&  0.621 &  0.561 &  0.475 & 0.408 \\
\#2&  0.618 &  0.560 &  0.478 & 0.411 \\
\#3&  \textbf{0.629} &  \textbf{0.570} &  \textbf{0.486} & \textbf{0.417} \\ 
\hline
\end{tabular}
\end{table}

In these prompts, \textit{text} represents the initial output generated by the LLM, which is the first prediction of the model and may contain errors that require correction. The \textit{candidate words} refers to a list of potential replacement words for the errors detected in {text}. These candidate words are obtained by breaking down the initial output text into individual characters and using a lookalike character word database to search for alternative words for each character that might have been predicted incorrectly. This ablation study reveals that the structure and phrasing of prompts play a crucial role in the model's text correction performance. As shown in Table~\ref{tab:prompt}, Prompt 3, with its concise, clear, and directive format, outperforms the other prompts across all BLEU scores. In contrast, prompts with more complex structures or redundant information may distract the model, leading to lower correction accuracy. These findings suggest that well-designed prompts with clear, focused instructions can significantly enhance the model’s correction capabilities.

\noindent $\bullet$ \textbf{Analysis of the Size of the Similar Word Database.~} 
In our framework, the effectiveness of the Glyph Error Correction Module heavily depends on the size and structure of the visually similar glyph database. To investigate how database size impacts recognition performance, we evaluate four configurations with varying maximum numbers of similar words for each glyph, i.e., 5, 7, 10, and 12 candidate words. The results, as shown in Table~\ref{tab:dict_size}, demonstrate the relationship between database size and BLEU scores.

From the results, we observe that increasing the number of candidate words from 5 to 7 provides a slight improvement in BLEU scores across all metrics. However, the most significant gain in performance is achieved when the database size is increased to 10 candidates per glyph, where the BLEU-1, BLEU-2, BLEU-3, and BLEU-4 scores reach their highest values. This suggests that a larger database, offering a broader range of correction options, significantly improves recognition accuracy, particularly for difficult or ambiguous characters.

However, when the database size is further increased to 12 candidates, the BLEU-2 score drops slightly, and other scores remain relatively stable. This indicates that beyond a certain point, increasing the number of candidate words may not yield additional improvements in accuracy. The reason for this decline could be that a larger database introduces more potential corrections, some of which may not be relevant or may lead to incorrect corrections due to ambiguity. This can confuse the model, especially when the visual similarity between candidate words is too high or when there is insufficient context to disambiguate between options, resulting in diminishing returns or even a reduction in performance.

\begin{table}[h]
\centering
\small 
\caption{Impact of the number of similar words in the database on recognition performance.}
\label{tab:dict_size}
\resizebox{\linewidth}{!}{ 
\begin{tabular}{c|c|c|c|c}
\hline
\textbf{Candidates} & \textbf{BLEU-1} & \textbf{BLEU-2} & \textbf{BLEU-3} & \textbf{BLEU-4} \\
\hline
5 & 0.611 & 0.551 & 0.470 & 0.403 \\
7 & 0.614 & 0.553 & 0.470 & 0.403 \\
10 & \textbf{0.629} & \textbf{0.570} & \textbf{0.486} & \textbf{0.417} \\
 12& 0.621& 0.563& 0.475&0.412\\ \hline
\end{tabular}} 
\end{table}

% \subsection{Efficiency Analysis} 

\subsection{Visualization} 
\begin{figure}
\centering
\includegraphics[width=\linewidth]{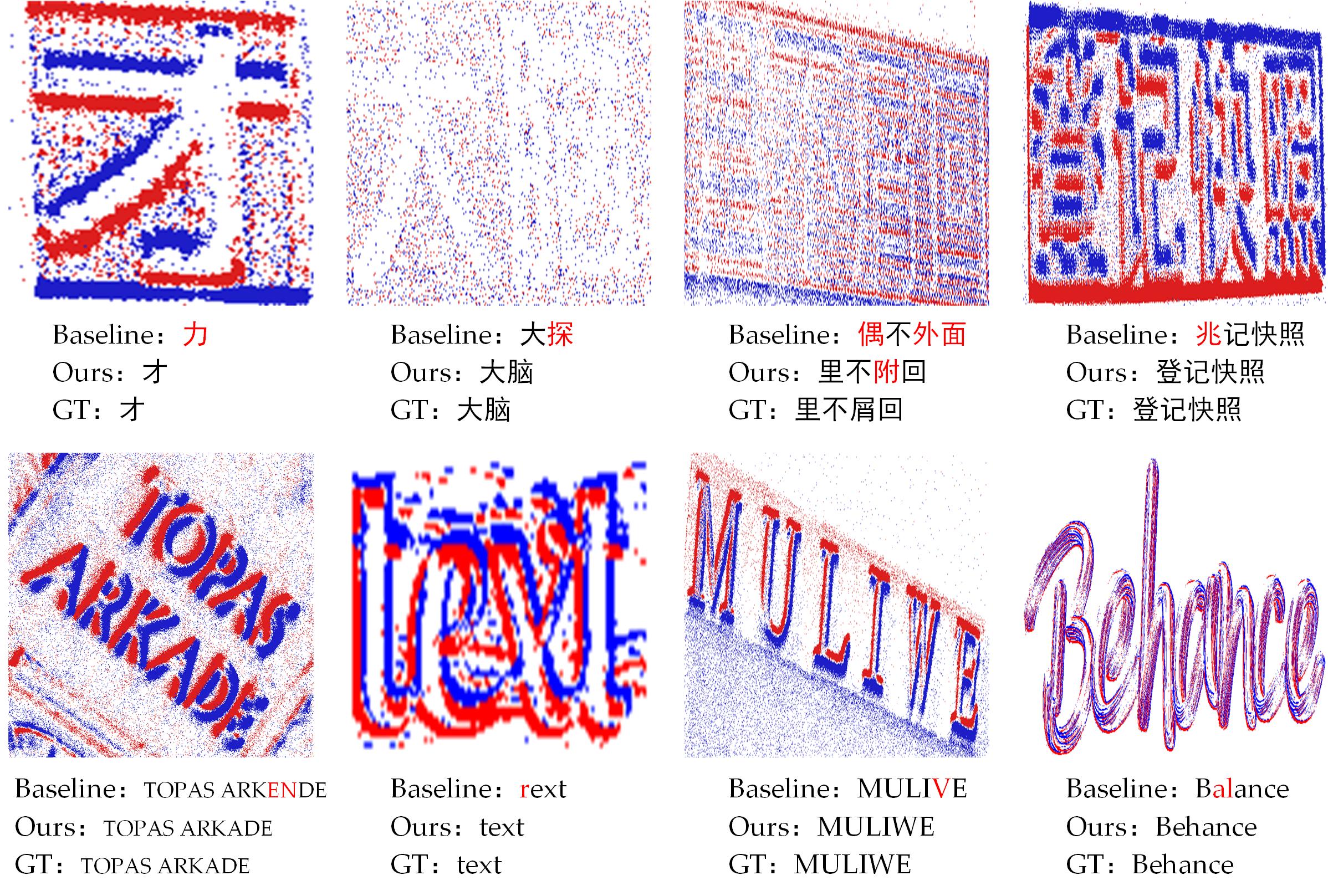}
\caption{Comparison of Baseline and Glyph-Corrected Recognition Results. Red text indicates misrecognized visually similar characters.}
\label{fig:correct}
\end{figure}

The Fig.~\ref{fig:correct} illustrates examples of successful text corrections achieved using our Glyph Error Correction Module. In each case, the baseline model output, our model’s corrected output, and the ground truth (GT) are presented for comparison. The visualizations highlight the module’s ability to improve text recognition, especially in challenging cases involving visually similar characters or complex text structures.

\begin{CJK*}{UTF8}{gbsn} 
In the first few examples, due to image blurring, visually similar glyph errors are observed. For instance, the baseline model misinterprets characters like ``才'' as ``力'' and ``里'' as ``偶''. In contrast, our module correctly identifies these characters using glyph-based corrections, aligning more accurately with the ground truth and demonstrating its effectiveness in disambiguating similar-looking characters.
\end{CJK*}

For English text, examples like ``rext'' and ``MULIVE'' show improvements where the baseline fails to capture certain letters accurately due to visual noise or distortions. For example, the baseline may recognize ``t'' as ``r'' or ``w'' as ``v'' due to similar shapes under noise. Our model’s output matches the intended text closely, indicating that the Glyph Error Correction Module successfully retrieves suitable alternatives from the similar words dictionary, leading to more precise text predictions. 

Overall, these visualizations confirm that the Glyph Error Correction Module enhances recognition accuracy by addressing both character-level and word-level errors, effectively correcting ambiguities in complex text.

\subsection{Limitation Analysis} 
Our EventSTR model faces two key limitations. First, it heavily relies on a large-scale pre-trained LLM, which demands significant computational resources, making it less suitable for real-time applications or deployment on resource-constrained devices. The high computational requirements can also result in slower inference times, posing challenges in efficiency-critical scenarios. Second, the model is initialized with weights pre-trained on Visual Question Answering (VQA) tasks, which, while effective for VQA, are not specifically optimized for text recognition tasks. This can lead to suboptimal performance in OCR scenarios, particularly when dealing with diverse and complex text layouts.

\section{Conclusion and Future Work} \label{sec::conclusion}
In this paper, we propose a novel event stream based scene text recognition task. A large-scale benchmark dataset is proposed for this research problem, termed EventSTR, which targets achieving high-performance and robust scene text recognition. The videos are collected using a high-definition Prophesee event camera and involve both Chinese and English text recognition. We also provide multiple baselines for this benchmark dataset and believe it will pave a new road for the event-based STR. In addition, we also propose a large language model based text recognition framework equipped with an error correction module and memory mechanism. Extensive experiments on multiple benchmark datasets fully validated the effectiveness of our proposed STR framework.

In future works, we will exploit new knowledge distillation strategies based on the SimC-ESTR framework to make it more lightweight and hardware-friendly. Also, the different efficient and low-latency event representations will also be an interesting research direction for the high-definition event-based STR task.

% \section*{Acknowledgment} 

\small{ 
\bibliographystyle{IEEEtran}
\bibliography{reference}
}

% that's all folks
\end{document}